\documentclass{CUP-JNL-EDS}

\usepackage{latexsym}
\usepackage{graphicx}
\usepackage{multicol,multirow}
\usepackage{amsmath,amssymb,amsfonts}
\usepackage{mathrsfs}
\usepackage{amsthm}
\usepackage{rotating}
\usepackage{appendix}
\usepackage[authoryear]{natbib}
\usepackage{ifpdf}
\usepackage[T1]{fontenc}
\usepackage{times}
\usepackage{newtxmath}
\usepackage{textcomp}%
\usepackage{xcolor}%
\usepackage{hyperref}
\usepackage{lipsum}

\jname{Environmental Data Science}
\jyear{2022}

\begin{document}

\begin{Frontmatter}

\title[Article Title]{\texttt{DL4DS} - Deep Learning for empirical DownScaling}

\author[1]{Carlos Alberto Gomez Gonzalez}\email{carlos.gomez@bsc.es}

\address[1]{\orgdiv{Earth Sciences department}, \orgname{Barcelona Supercomputing Center}, \orgaddress{\city{Barcelona}, \state{Catalonia},  \country{Spain}}}


\received{28 January 2022}

\authormark{Gomez Gonzalez}



\abstract{A common task in Earth Sciences is to infer climate information at local and regional scales from global climate models. Dynamical downscaling requires running expensive numerical models at high resolution which can be prohibitive due to long model runtimes. On the other hand, statistical downscaling techniques present an alternative approach for learning links between the large- and local-scale climate in a more efficient way. A large number of deep neural network-based approaches for statistical downscaling have been proposed in recent years, mostly based on convolutional architectures developed for computer vision and super-resolution tasks. This paper presents \texttt{DL4DS}, Deep Learning for empirical DownScaling, a python library that implements a wide variety of state-of-the-art and novel algorithms for downscaling gridded Earth Science data with deep neural networks. \texttt{DL4DS} has been designed with the goal of providing a general framework for training convolutional neural networks with configurable architectures and learning strategies to facilitate the conduction of comparative and ablation studies in a robust way. We showcase the capabilities of \texttt{DL4DS} on air quality CAMS data over the western Mediterranean area. The \texttt{DL4DS} library can be found in this repository: \url{https://github.com/carlos-gg/dl4ds}}

\policy{This paper presents \texttt{DL4DS} the first open-source library with state-of-the-art and novel deep learning algorithms for empirical downscaling.}
\end{Frontmatter}

\section{Introduction}

Downscaling aims to bridge the gap between the large spatial scales represented by global climate models (GCMs) to the smaller scales required for assessing regional climate change and its impacts \citep{maraun17}. This task can be approached via dynamical or statistical downscaling. In the former, a high-resolution regional climate model is nested into a GCM over the domain of interest \citep{rummukainen10}, while the latter aims to learn empirical links between the large- and local-scale climate that are in turn applied to low-resolution climate model output. Statistical or empirical downscaling comes with the benefit of a lower computational cost compared to its dynamical counterpart. 

Machine Learning (ML) can be defined as the study of computer algorithms that improve automatically by finding statistical structure (building a model) from training data for automating a given task. The field of ML started to flourish in the 1990s and has quickly become the most popular and most successful subfield of Artificial Intelligence (AI) thanks to the availability of faster hardware and larger training datasets. Deep Learning \citep[DL,][]{bengio21} is a specific subfield of ML aimed at learning representations from data putting emphasis on learning successive layers of increasingly meaningful representations \citep{dlwp21}. DL has shown potential in a wide variety of problems in Earth sciences dealing with high-dimensional and complex data and offers exciting new opportunities for expanding our knowledge about the Earth system \citep{reich19, huntin19, dewitte21, irrgang21}.

The layered representations in DL are learned via models called neural networks, where an input n-dimensional tensor is received and multiplied with a weights tensor to produce an output. A bias term is usually added to this output before passing the result through an element-wise nonlinear function (also called activation). Convolutional Neural Networks \citep[CNNs,][]{cnns_lenet} are a type of neural network that rely on the convolution operation, a linear transformation that takes advantage of the implicit structure of gridded data. The convolution operation uses a weight tensor that operates in a sliding window fashion on the data, meaning that only a few input grid points contribute to a given output and that the weights are reused since they are applied to multiple locations in the input. CNNs have been used almost universally for the past few years in computer vision applications, such as object detection, semantic segmentation and super-resolution. CNNs show performances on par with more recent models such as visual Transformers \citep{convnext22} while being more efficient in terms of memory and computations.

In spite of recent efforts in the computer science community, the field of DL still lacks solid mathematical foundations. It remains fundamentally an engineering discipline heavily reliant on experimentation, empirical findings and software developments. While scientific software tools such as \texttt{Numpy} \citep{numpy20}, \texttt{Xarray} \citep{xarray17} or \texttt{Jupyter} \citep{jupyter} have an essential role in modern Earth sciences research workflows, state-of-the-art domain-specific DL-based algorithms are usually developed as proof-of-concept scripts. For DL to fulfill its potential to advance Earth sciences, the development of AI- and DL-powered scientific software must be carried out in a collaborative and robust way following open-source and modern software development principles. 

\section{CNN-based super-resolution for statistical downscaling}

Statistical downscaling of gridded climate variables is a task closely related to that of super-resolution in computer vision, considering that both aim to learn a mapping between low-resolution and high-resolution grids \citep{sr_wang21}. Unsurprisingly, several DL-based approaches have been proposed for statistical or empirical downscaling of climate data in recent years \citep{vandal17, leinonen20, stengel20, hohlein20, liu20, harilal21}. Most of these methods have in common the use of convolutions for the exploitation of multivariate spatial or spatio-temporal gridded data, that is 3D (\texttt{height/latitude}, \texttt{width/longitude}, \texttt{channel/variable}) or 4D (\texttt{time}, \texttt{height/latitude}, \texttt{width/longitude}, \texttt{channel/variable}) tensors. Therefore, CNNs are efficient at leveraging the high-resolution signal from heterogeneous observational datasets, from now on called predictors or auxiliary variables, while downscaling low-resolution climatological fields. 

In our search for efficient architectures for empirical downscaling, we have developed \texttt{DL4DS}, a library that draws from recent developments in the field of computer vision for tasks such as image-to-image translation and super-resolution. \texttt{DL4DS} is implemented in \texttt{Tensorflow/Keras}, a popular DL framework, and contains a collection of building blocks that abstract and modularize a few key design strategies for composing and training empirical downscaling DL models. These strategies are discussed in the following section.

\section{\texttt{DL4DS} core design principles and building blocks} \label{sec:coredesign}

The core design principles and general architecture of \texttt{DL4DS} are discussed in the following subsections. An overall architecture of \texttt{DL4DS} is shown in Fig. \ref{fig:workflow}, while some of the building blocks and network architectures are shown in Figs. \ref{fig:blocks} and \ref{fig:archs}.

\subsection{Type of statistical downscaling}
\begin{figure}
\centering
\includegraphics[width=\columnwidth]{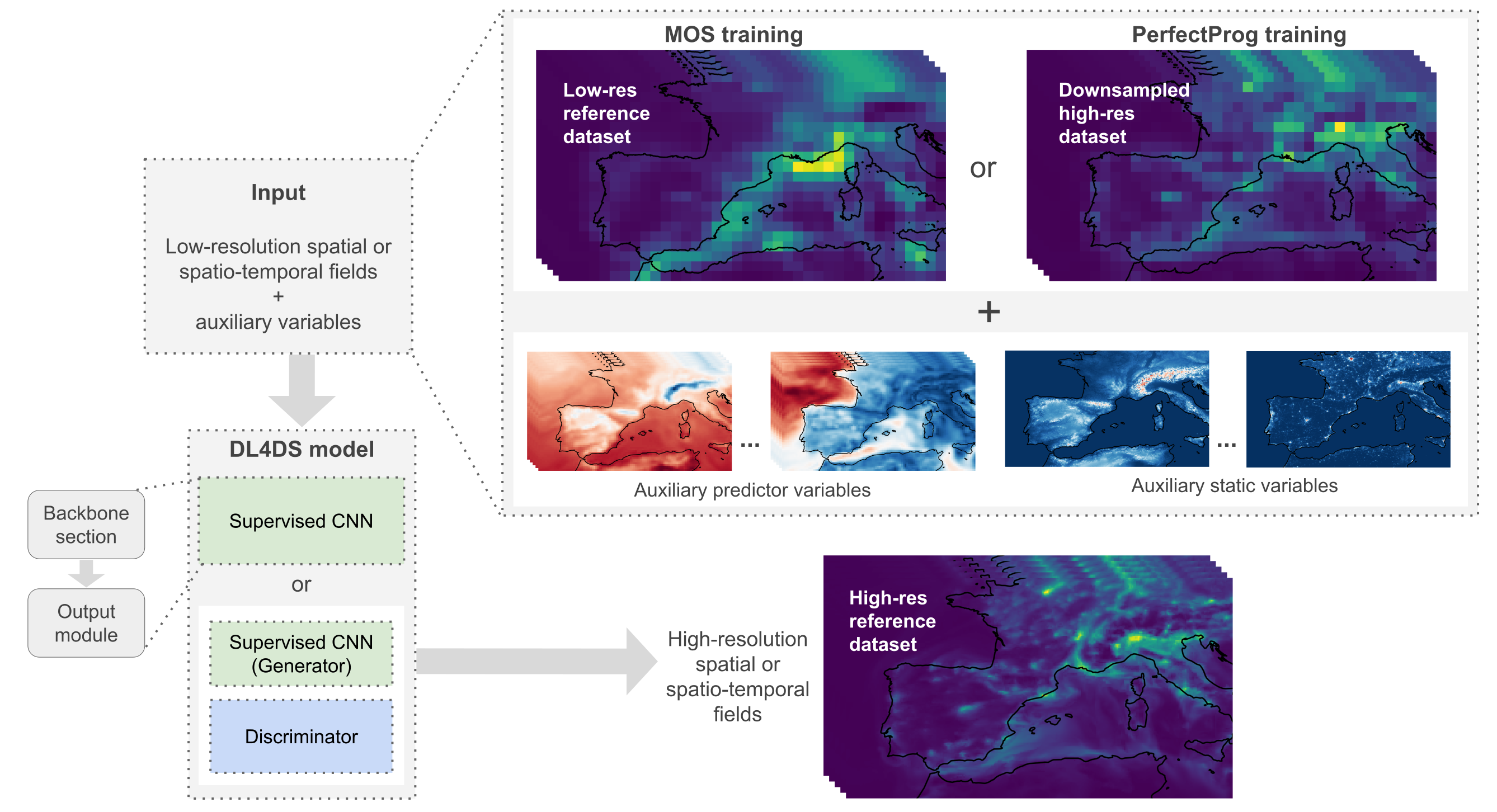}
\caption{The general architecture of \texttt{DL4DS}. A low-resolution gridded dataset can be downscaled, with the help of auxiliary predictor and static variables, and a high-resolution reference dataset. The mapping between the low- and high-resolution data is learned with either a supervised or a conditional generative adversarial DL model}
\label{fig:workflow}
\vspace{-4.mm}
\end{figure}

Statistical downscaling methods aim to derive empirical relationships between an observed high-resolution variable or predictand and low-resolution predictor variables \citep{maraun17}. Two main types of statistical downscaling can be defined depending on the origin of the predictors used for training; Model output statistics (MOS), where the predictors are taken directly from global climate model outputs, and Perfect Prognosis (PerfectProg) methods, that relies on observational datasets for both predictand and predictors (see Fig. \ref{fig:workflow}). Most of the DL-based downscaling methods proposed to date, work in PerfectProg setup, where a high-resolution observational dataset is used to create paired training samples via a downsampling or coarsening operation \citep{vandal17, leinonen20, stengel20}. The trained model is then applied to the desired, unseen during training, low-resolution data in a domain-transfer fashion. Other approaches such as those proposed by \cite{hohlein20} or \cite{harilal21}, model a cross-scale transfer function between explicit low-resolution and high-resolution datasets. \texttt{DL4DS} supports both training frameworks, with explicit paired samples (in MOS fashion) or with paired samples simulated from high-resolution fields (in PerfectProg fashion).

\subsection{Multivariate modeling and auxiliary variables}
Using high-resolution static predictors fields, such as, topography, distance from the sea, and land-sea mask, as well as high- or intermediate-resolution essential climate variables, can help improve the inference capacity of empirical downscaling methods \citep{maraun17, vandal17, hohlein20}. The most straightforward way to include these additional predictors and static variables is to merge (concatenate) them across the channel dimension with the input low-resolution fields. \texttt{DL4DS} can handle an arbitrary number of time-varying high- or intermediate-resolution predictors and high-resolution static variables. Additionally, static high-resolution fields are passed through a convolutional block in the output module as shown in panel (E) of Fig. \ref{fig:archs} in order to emphasize high-resolution topographic information.

\subsection{Pre-processing and normalization}
\texttt{DL4DS} implements a couple of pre-processing methods aimed at normalizing or standardizing multi-dimensional gridded datasets, a required step when splitting the data into train and validation/test sets used for training and testing DL models. These procedures, namely mix-max and standardization scalers, are implemented as classes (\texttt{dl4ds.MinMaxScaler} and \texttt{dl4ds.StandardScaler}) in the preprocessing module of \texttt{DL4DS}. These classes extend the functionality of methods from the \texttt{Scikit-learn} package \citep{scikit-learn} aimed at scaling tabular or 2D data. \texttt{DL4DS} scalers can handle 3D and 4D datasets with NaN values in either \texttt{numpy.ndarray} or \texttt{xarray.Dataarray} formats. 

\begin{figure}[ht]
\centering
\includegraphics[width=\columnwidth]{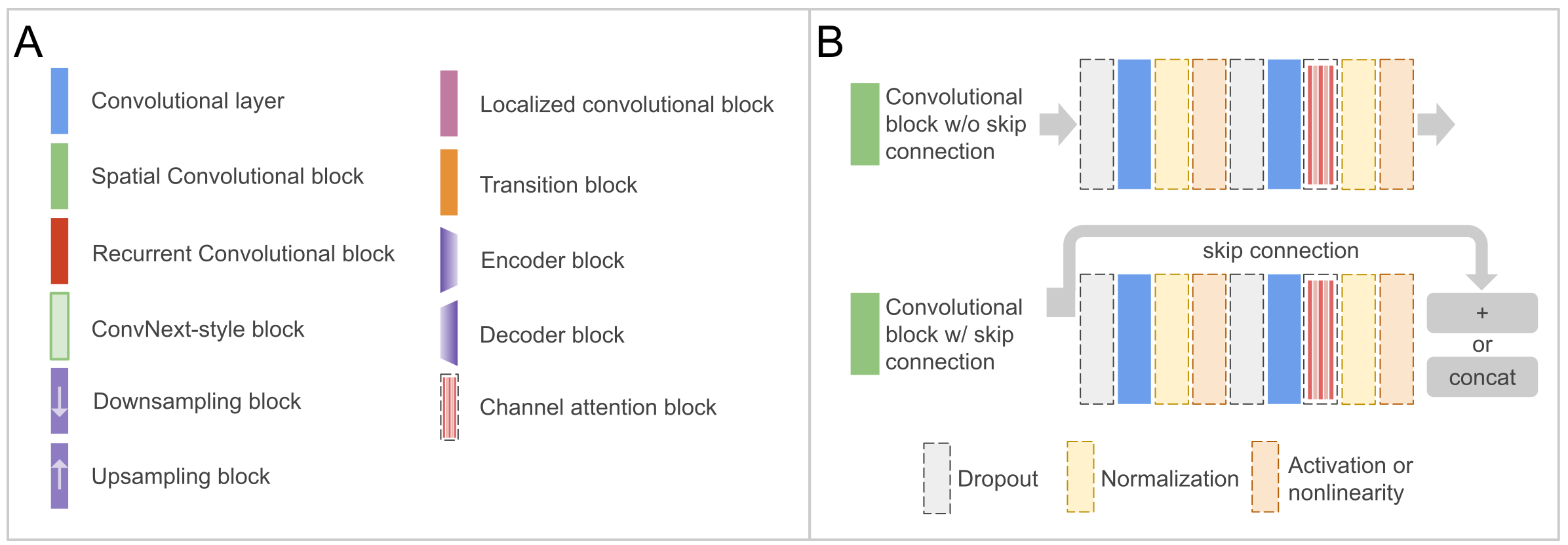}
\caption{Panel (A) shows the main blocks and layers implemented in \texttt{DL4DS}. Panel (B) shows the structure of the main spatial convolutional block, a succession of two convolutional layers with interleaved regularization operations, such as dropout or normalization. Blocks and operations shown with dashed lines are optional}
\label{fig:blocks}
\vspace{-4.mm}
\end{figure}

\subsection{Architecture of the super-resolution models in \texttt{DL4DS}}
The super-resolution networks implemented in \texttt{DL4DS} aim at learning a mapping from low or coarse-resolution to high-resolution grids. The super-resolution networks in \texttt{DL4DS} are composed of two main parts, a backbone section and an output module, as depicted in Fig. \ref{fig:workflow}.

\subsubsection{Backbone section} \label{sec:backbone}
\begin{figure}[ht]
\centering
\includegraphics[width=\columnwidth]{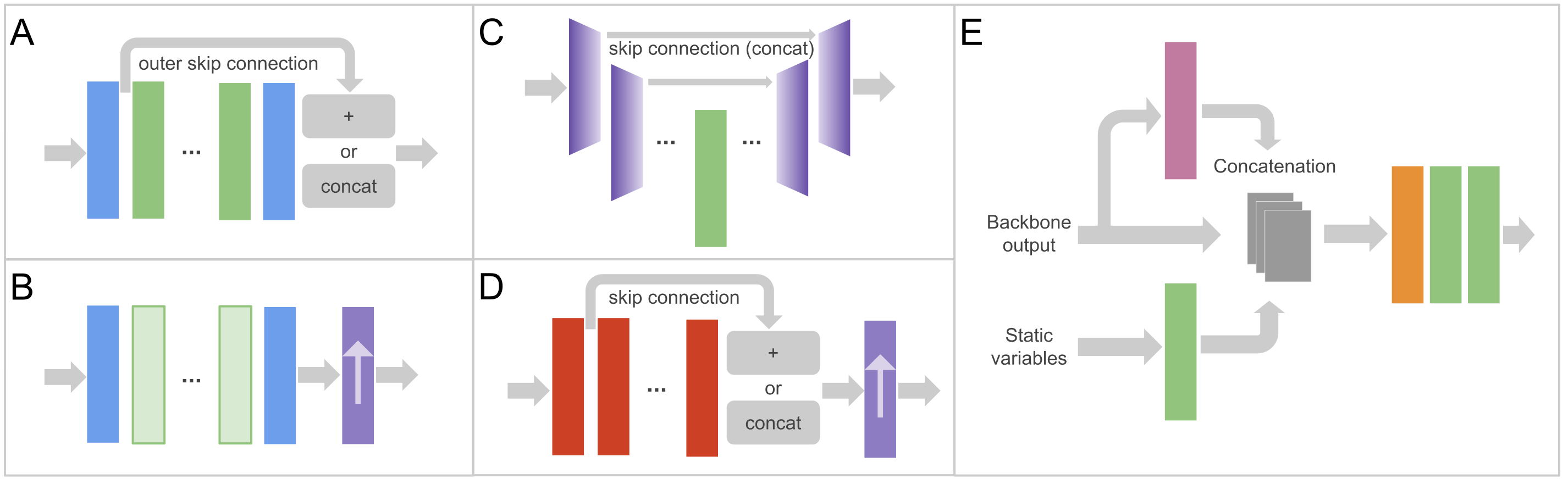}
\caption{\texttt{DL4DS} supervised DL models, as well as generators, are composed of a backbone section (examples in panels (A), (B), (C) and (D)) and an output module (panel (E). Panel (A) shows the backbone of models for downscaling pre-upsampled spatial samples using either residual or dense blocks. Panel (B) presents the backbone of a model for downscaling spatial samples using ConvNext-like blocks and one of the post-upsampling blocks described in subsection \ref{ssec:upsampling}. Panel (C) shows the backbone of a model for downscaling pre-upsampled spatial samples using an encoder-decoder structure. Panel (D) shows the backbone of a model for downscaling spatio-temporal samples using recurrent-convolutional blocks and a post-upsampling block. These backbones are followed by the output module (see subsection \ref{ssec:outmod}) shown in panel (E). The color legend for the blocks used here is shown in panel (A) of Fig. \ref{fig:blocks}}
\label{fig:archs}
\end{figure}

We adopt several strategies for the design of the backbone sections, as shown in panels (A) through (D) of Fig. \ref{fig:archs}. The main difference being the particular arrangement of convolutional layers. In \texttt{DL4DS}, the backbone type can be set with the \texttt{backbone} parameter of the training classes mentioned subsection \ref{ssec:losses}:

\begin{itemize}
\item \texttt{Convnet} backbone - Composed of standard convolutional blocks. As shown in panel (B) of Fig. \ref{fig:blocks}, each spatial convolutional block is composed of two convolutional layers (using 3x3 kernels) with some interleaved optional operations, such as dropout \citep{dropout}, batch/layer normalization or a channel attention block (discussed in subsection \ref{ssec:attention}).
\item \texttt{Resnet} backbone - Composed of residual blocks or convolutional blocks with skip connections\footnote{Skip or shortcut connections are connections that feed the output of a particular layer to later non-sequential or non-adjacent layers in the network. Element-wise addition is used in residual blocks, while concatenation is used in dense blocks or in decoder blocks of a U-Net.} that learn residual functions with reference to the layer inputs. Residual skip connections have been widely employed in super-resolution models \citep{sr_wang21} in the past. In \texttt{DL4DS}, we implemented cross-layer skip connections within each residual block and outer skip connections at the backbone section level \citep{resnet}.
\item \texttt{Densenet} backbone - Composed of dense blocks (see panel (B) of Fig. \ref{fig:blocks}). As in the case of the \textit{Resnet} backbone, we implement cross-layer skip connections within each dense block and outer skip connections at the backbone section level \citep{densenet}.
\item \texttt{Unet} backbone - Inspired by the structure of the U-Net DL model \citep{unet}, this backbone is composed of encoder and decoder blocks with skip connections, as depicted in panel (C) of Fig. \ref{fig:archs}. The encoder block consists of a convolutional block plus downsampling via max pooling\footnote{Max pooling is a sample-based discretization process that selects the maximum element from the region of the feature map covered by the filter.}, while the decoder block consists of upsampling, concatenation with skip feature channels\footnote{A feature channel or feature map refers to the output of a convolutional layer.} from the encoder section, followed by a convolutional block. This backbone is only implemented for pre-upsampled spatial samples. 
\item \texttt{Convnext} backbone - Based on the recently proposed ConvNext model \citep{convnext22}, we implemented a backbone with depth-wise convolutional layers and larger kernel sizes (7x7). To the best of our knowledge, this is the first time that a ConvNext-style block architecture has been applied to the task of statistical downscaling.
\end{itemize}
 
\paragraph{Spatial and spatio-temporal modelling} \label{ssec:spsptsamples}
While CNNs excel at modelling spatial or gridded data, hybrid convolutional and recurrent architectures are designed to exploit spatio-temporal sequences. Convolutional Long Short-Term Memory \citep[ConvLSTM,][]{convlstm15} and convolutional Gated-Recurrent-Units \citep{convgru15} are examples of models that have been used for downscaling time-evolving gridded data \citep{leinonen20, harilal21}. \texttt{DL4DS} can model either spatial or spatio-temporal data, by either using standard convolutional blocks or recurrent convolutional blocks. The spatio-temporal network in panel (D) of Fig. \ref{fig:archs} contains recurrent convolutional blocks to handle 4D training samples (with time dimension). The structure of these recurrent convolutional blocks is similar to that of the main convolutional block shown in panel (B) of Fig. \ref{fig:blocks} but using convolutional LSTM layers instead of standard convolutional ones. 

\paragraph{Channel attention} \label{ssec:attention}
In \texttt{DL4DS}, we implement a channel attention mechanism based on those of the Squeeze-and-Excitation networks \citep{squeeze-and-excitation} and the Convolutional Block Attention Module \citep{cbam_woo18}. This attention mechanism exploits the inter-channel relationship of features by providing a weight for each channel in order to enhance those that contribute the most to the optimization and learning process. First, it aggregates spatial information of a feature map by using average pooling\footnote{Average pooling is a sample-based discretization process that computes the average value from the region of the feature map covered by the filter.}. The resulting vector is then passed through two 1x1 convolutional blocks and a sigmoid activation to create the channel weights (attention maps) which are multiplied element-wise to each corresponding feature map. The channel attention mechanism is integrated as an optional step in \texttt{DL4DS} convolutional blocks to get refined feature maps throughout the network. 

\paragraph{Upsampling techniques} \label{ssec:upsampling}
The upsampling operation is a crucial step for DL-based empirical downscaling methods. In computer vision, upsampling refers to increasing the number of rows and columns, and therefore the number of pixels, of an image. Increasing the size of an image is also often called upscaling, which oddly enough carries the opposite meaning in the weather and climate jargon. When it comes to downscaling or super-resolving gridded data, it is of utmost importance to preserve fine-scale information, or to transfer it from the high-resolution to the low-resolution fields, while increasing the number or grid points in both horizontal and vertical directions. In \texttt{DL4DS}, we implement several well-established upsampling methods proposed in the super-resolution literature \citep{sr_wang21}. These methods belong to two main upsampling strategies, depending on whether the upsampling happens before the network or inside of it: pre-upsampling and post-upsampling. In the former case, the models are fed with a pre-upsampled (via interpolation) low-resolution input and a reference high-resolution predictand. In the latter case, the low-resolution input is directly fed together with the high-resolution predictand. In this scenario, we learn a mapping from low-resolution to high-resolution fields in low-dimensional space with end-to-end learnable layers integrated at the end of the backbone section (see panels (B) and (D) of Fig. \ref{fig:archs}). This results in an increased number of learnable parameters, but an increased computational efficiency thanks to working on smaller grids. When working with \texttt{DL4DS}, the upsampling method can be set with the \texttt{upsampling} parameter of the training classes discussed in the subsection \ref{ssec:losses}: 

\begin{itemize}
\item \texttt{PIN} pre-upsampling - Models trained with interpolation-based pre-upsampled input.
\item \texttt{RC} post-upsampling - Models including a resize convolution block (bilinear interpolation followed by a convolutional layer).
\item \texttt{DC} post-upsampling - Models including a deconvolution (also called transposed convolution) block \citep{deconvolution}. With a transposed convolution we perform a transformation opposite to that of a normal convolution. In practice, the image is expanded by inserting zeros and performing a convolution. This upsampling method can sometimes produce checkerboard-like patterns that hurt the quality of the downscaled or super-resolved products.
\item \texttt{SPC} post-upsampling - Models including a subpixel convolution block \citep{edsr_spc17}. A subpixel convolution consists of a regular convolution with $s^2$ filters (where $s$ is the scaling factor) followed by an image reshaping operation called a phase shift. Instead of putting zeros in between grid points as it is done in a deconvolution layer, we calculate more convolutional filters in lower resolution and reshape them into an upsampled grid. 
\end{itemize}

In the context of empirical downscaling, pre-upsampling has been used by \citet{vandal17, harilal21}, resize convolution by \cite{leinonen20}, transposed convolution by \cite{hohlein20} and sub-pixel convolution by \cite{liu20}. Choosing an upsampling method depends on the problem at hand (MOS or PerfectProg training) and the computational resources available.

\subsubsection{Output module} \label{ssec:outmod}
The second part of every \texttt{DL4DS} network is the output module, as shown in panel (E) of Fig. \ref{fig:archs}. Here we concatenate the intermediate outputs of three branches: the backbone section feature channels, the backbone section output passed through a localized convolutional block, and a separate convolutional block over the input high-resolution static variables. These concatenated feature channels are in turn passed through a transition block, a 1x1 convolution used to control the number of channels, and two final convolutional blocks. The first of this two convolutional blocks in the output module applies a channel attention block by default. 

\paragraph{Localized convolutional block} \label{ssec:lws}
CNNs bring many positive qualities for the exploitation of spatial or gridded data. However, the property of translation invariance of CNNs is usually sub-optimal in geospatial applications where location-specific information and dynamics need to be taken into account. This has been addressed by \cite{uselis20} for the task of weather forecasting. Within \texttt{DL4DS}, we implement a Localized Convolutional Block (LCB) located in the output module of our networks. This LCB consists of a bottleneck transition layer, via 1x1 convolutions to compresse the number of incoming feature channels, followed by a locally connected layer with biases (similar to a convolutional one, except that the weights here are not shared). To the best of our knowledge, this is the first time the idea of LCBs has been applied to the task of statistical downscaling.  

\subsection{Loss functions} \label{ssec:losses}
Loss functions are used to measure reconstruction error while guiding the optimization of neural networks during the training phase. 

\subsubsection{Supervised losses}
\texttt{DL4DS} implements pixel-wise losses, such as the Mean Squared Error (MSE or L2 loss) or the Mean Absolute Error (MAE or L1 loss), as well as structural dissimilarity (DSSIM) and multi-scale structural dissimilarity losses (MS-DSSIM), which are derived from the Structural Similarity Index Measure \citep[SSIM,][]{ssim}. The SSIM is a computer vision metric used for measuring the similarity between two images, based on three comparison measurements: luminance, contrast and structure. The MS-DSSIM loss has been used by \cite{chaudhuri20} for the task of statistical downscaling. 

\subsubsection{Adversarial losses}
\begin{figure}[ht]
\centering
\includegraphics[width=\columnwidth]{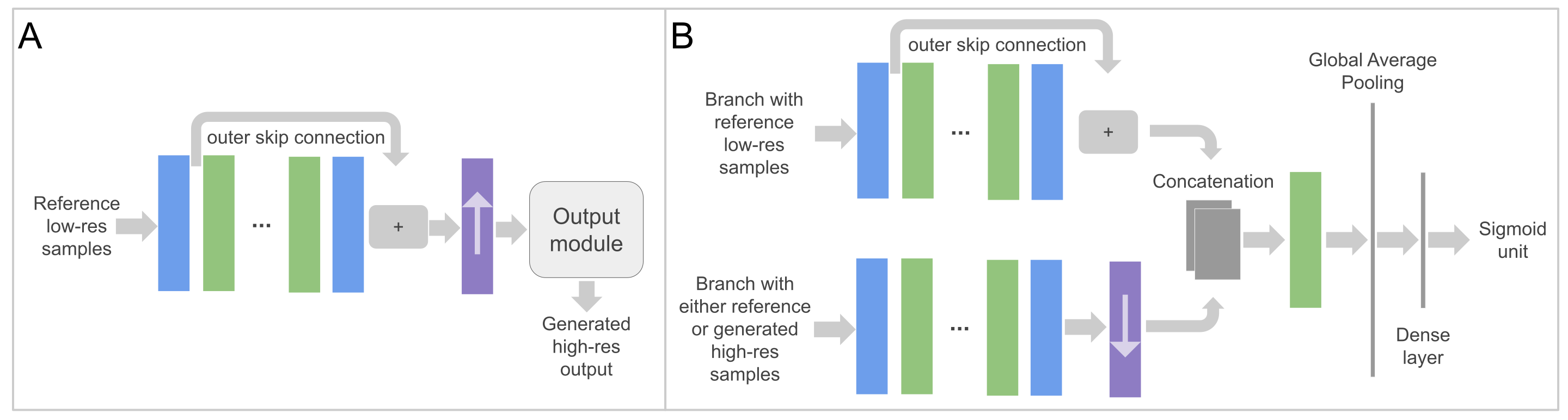}
\caption{Example of a conditional generative adversarial model for spatio-temporal samples in post-upsampling mode (see subsection \ref{sec:backbone}). Two networks, the generator shown in panel (A), and discriminator shown in panel (B), are trained together optimizing an adversarial loss (see subsection \ref{ssec:losses}). The color legend for the blocks used here is shown in panel (A) of Fig. \ref{fig:blocks}}
\label{fig:gan}
\end{figure}

Generative adversarial networks \citep[GANs,][]{gans} are a class of ML models in which two neural networks contest with each other in a zero-sum game, where one agent's gain is another agent's loss. This adversarial training was extended for its use with images for image-to-image translation problems by \cite{isola17}. Conditional GANs map input to output images while learning a loss function to train this mapping, usually allowing a better reconstruction of high-frequency fine details. \texttt{DL4DS} implements a conditional adversarial loss \citep{isola17} by training the two networks depicted in Fig. \ref{fig:gan}. The role of the generator is to learn to generate high-resolution fields from their low-resolution counterparts while the discriminator learns to distinguish synthetic high-resolution fields from reference high-resolution ones. Through iterative adversarial training, the resulting generator can produce outputs consistent with the distribution of real data, while the discriminator cannot distinguish between the generated high-resolution data and the ground truth. Adversarial training has been used in DL-based downscaling approaches proposed by \cite{leinonen20}, \cite{stengel20} and \cite{chaudhuri20}. In out tests, the generated high-resolution fields produced by the CGAN generators exhibit moderate variance, even when using the Monte Carlo dropout technique \citep{mcdropout} which amounts to applying dropout at inference time. 

When working with \texttt{DL4DS}, the training strategy is controlled by choosing one of the training classes: \texttt{dl4ds.SupervisedTrainer} or \texttt{dl4ds.CGANTrainer}, for using supervised or adversarial losses accordingly.

\section{Experimental results} \label{sec:results}
\begin{figure}
\centering
\includegraphics[width=\columnwidth]{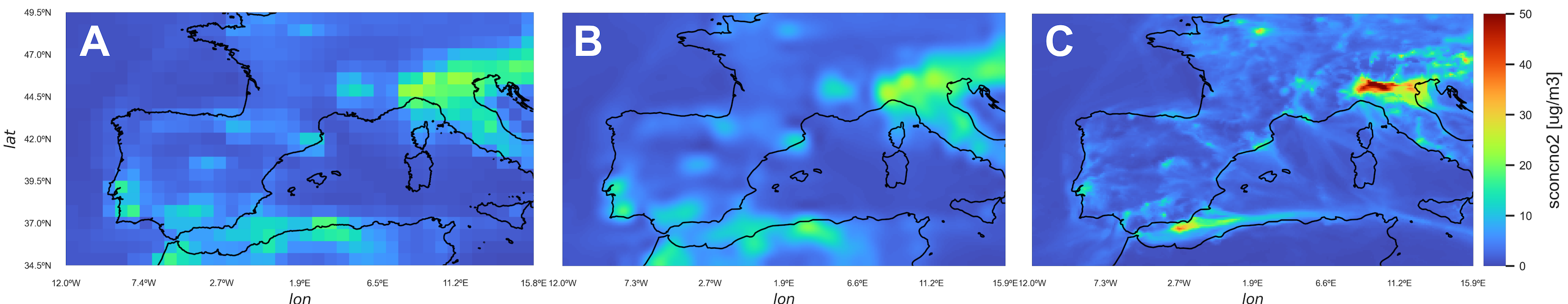}
\caption{A reference NO2 surface concentration field from the low-resolution CAMS global reanalysis is shown in panel (A). In panel (B), we present a resampled version, via bicubic interpolation, of the low-resolution reference field. This interpolated field looks overly smoothed and showcases the inefficiency of simple resampling methods at restoring fine scale information. Panel (C): corresponding high-resolution field from the CAMS regional reanalysis. Both low- and high-resolution grids were taken from the holdout set for the same time step. The maximum value shown corresponds to the maximum value in the high-resolution grid}
\label{fig:plots_ref}
\end{figure}

Downscaling applications are highly dependant on the target variable/dataset at hand, and therefore the optimal model architecture should be found in a case-by-case basis. In this section, we give a taste of the capabilities of \texttt{DL4DS} without conducting a rigorous comparison of model architectures and learning configurations. For our tests, we use data from the Copernicus Atmosphere Monitoring Service (CAMS) reanalysis \citep{camsra}, the latest global reanalysis dataset of atmospheric composition produced by the European Centre for Medium-Range Weather Forecasts (ECMWF), which provides estimates of Nitrogen dioxide (NO2) surface concentration. We select NO2 data from the period between 2014 and 2018 at a 3-hourly temporal resolution which results in $\sim$14.6k temporal samples. For our low-resolution dataset, we use the CAMS global reanalysis (CAMSRA) with an horizontal resolution of $\sim$80 km. Our high-resolution target is the CAMS regional reanalysis produced for the European domain with a spatial resolution of $\sim$10km (0.1º). We also include predictor atmospheric variables from the ECMWF ERA5 reanalysis \citep{era520}, namely 2 meter temperature and 10 meter wind speed, with an intermediate resolution of $\sim$25 km (0.25º). Finally, we include as static variables: topography from the Global Land One-km Base Elevation Project and a land-ocean mask derived from it, and a layer with the urban area fraction derived from the land cover dataset produced by the European Space Agency. In order to make our test less heavy on memory requirements, we spatially subset the data and focus on the western Mediterranean region, as shown in Fig. \ref{fig:plots_ref}.

\begin{table}[ht]
\caption{\texttt{DL4DS} models showcased in section \ref{sec:results}. The column named `panel' points the corresponding plot in Fig. \ref{fig:plots_samples}, Fig. \ref{fig:plots_corr} and Fig. \ref{fig:plots_rmse} \label{tab:models}.}
\centering
 \begin{tabular}{c c c c c c c c} 
 \hline
 Panel & Downscaling & Learning & Sample type & Backbone & Upsampling & LCB & loss \\ [0.5ex] 
 \hline\hline
 A & PerfectProg    & supervised    & spatial           & unet      & PIN & no  & mae \\ 
 B & MOS            & supervised    & spatial           & unet      & PIN & no  & mae \\
 C & PerfectProg    & supervised    & spatial           & resnet    & SPC & no  & dssim+mae \\
 D & MOS            & supervised    & spatial           & resnet    & SPC & yes & dssim+mae \\
 E & MOS            & supervised    & spatial           & resnet    & SPC & yes & mae  \\
 F & MOS            & adversarial   & spatial           & resnet    & SPC & yes & mae \\
 G & MOS            & supervised    & spatiotemp        & resnet    & SPC & yes & dssim+mae \\
 H & MOS            & supervised    & spatial           & convnext  & SPC & yes & mae \\ [1ex]
 \hline
 \end{tabular}
\end{table}

We showcase eight models trained with \texttt{DL4DS}, without the intention of a full exploration of possible architectures and learning strategies. Different loss functions, backbones, learning strategies and other parameters are combined in the model architectures detailed in Table \ref{tab:models}. The training is carried out using a single cluster node with four NVIDIA V100 GPUs. The data is split into train and test sets by reserving the last year of data for test and validation. We fit the \texttt{dl4ds.StandardScaler} on the training set (getting the global mean and standard deviation) and applied it to both training and test sets (subtracting the global mean and dividing by the global standard deviation). All models are trained with the Adam optimizer \citep{adam}, for 100 epochs in the case of supervised CNNs and 18 epochs in the case of the conditional adversarial models.

\begin{figure}[ht]
\centering
\includegraphics[width=\columnwidth]{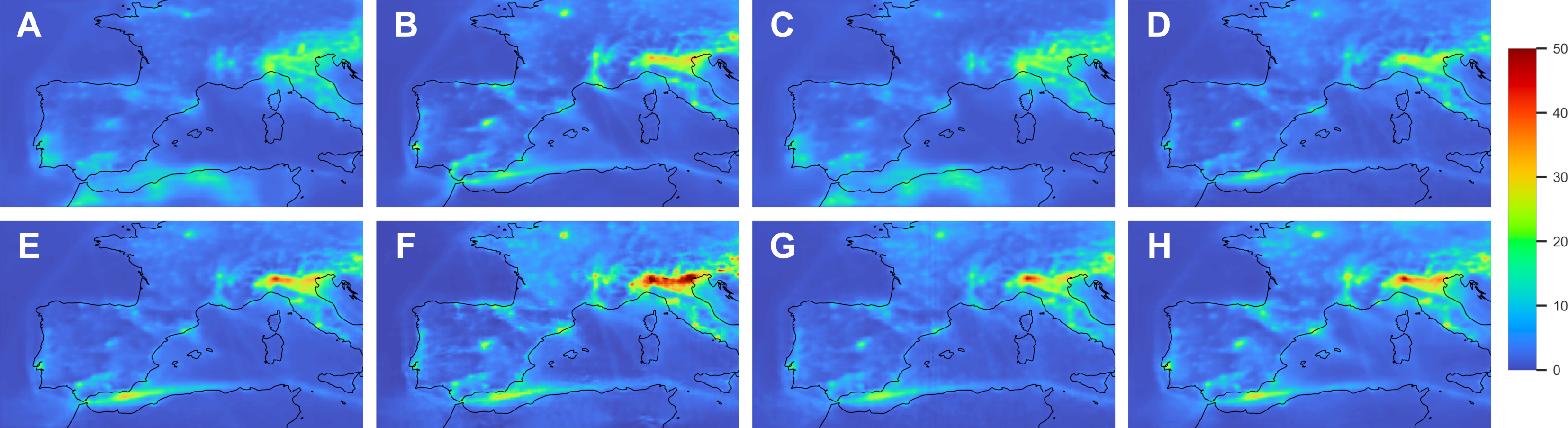}
\caption{Examples of downscaled products obtained with \texttt{DL4DS}, corresponding to the reference grid shown in panel (A) of Fig. \ref{fig:plots_ref}. The models corresponding to each panel are detailed in Table \ref{tab:models}}
\label{fig:plots_samples}
\end{figure}

\begin{figure}[ht]
\centering
\includegraphics[width=\columnwidth]{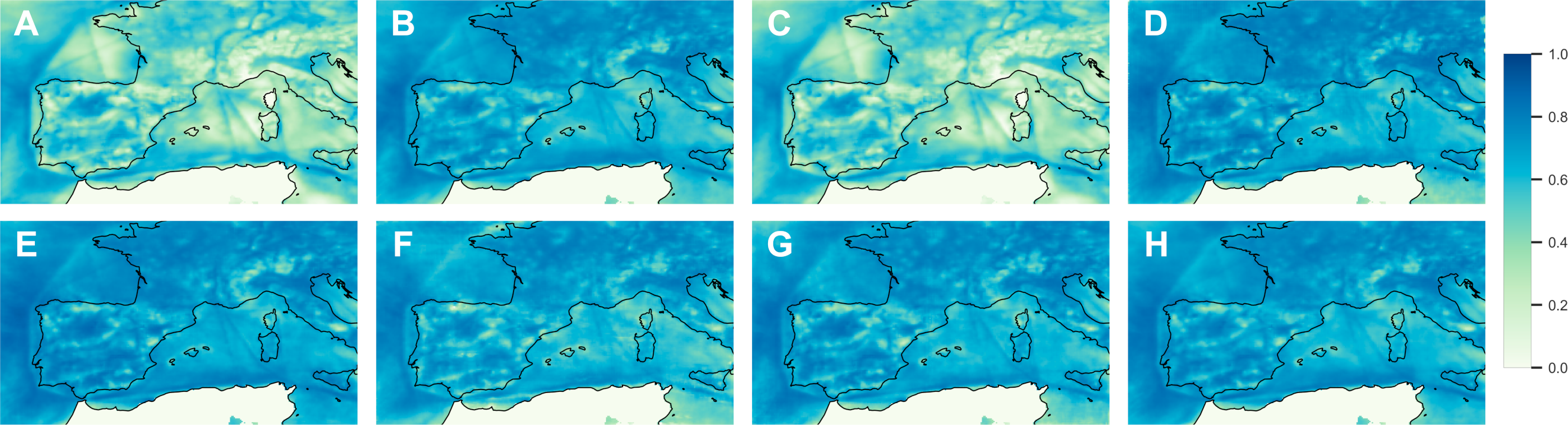}
\caption{Pixel-wise Pearson correlation for each model, computed for the whole year of 2018}
\label{fig:plots_corr}
\end{figure}

\begin{figure}[ht]
\centering
\includegraphics[width=\columnwidth]{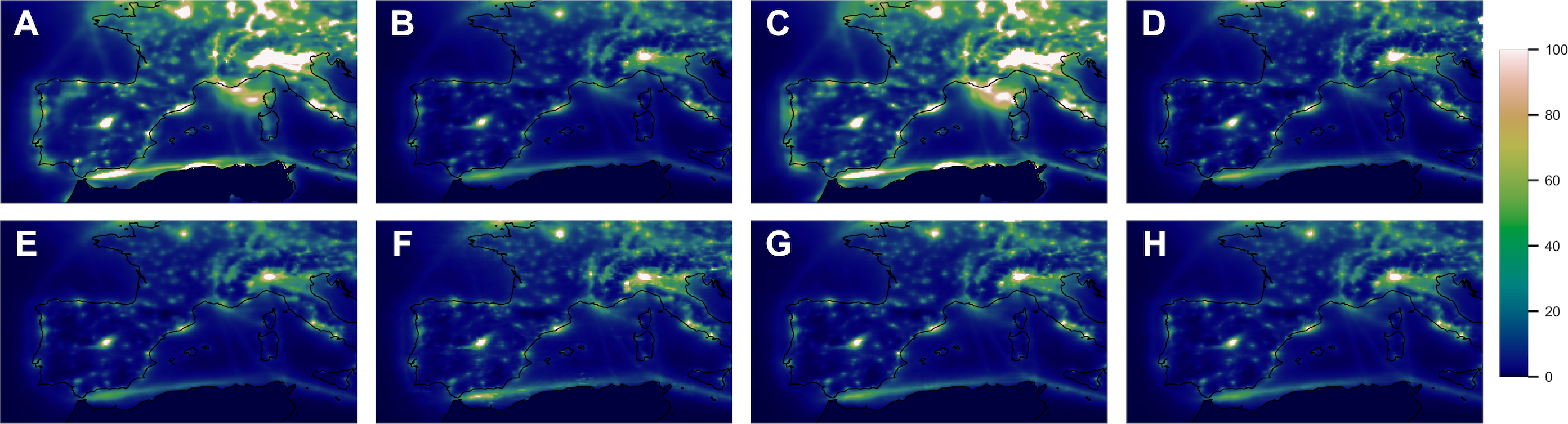}
\caption{Pixel-wise RMSE for each model, computed for the whole year of 2018. The dynamic range is shared for all the panels, with a fixed maximum value to facilitate the visual comparison}
\label{fig:plots_rmse}
\end{figure}

In Fig. \ref{fig:plots_samples}, we show examples of predicted high-resolution fields produced by the models detailed in Table \ref{tab:models} and corresponding to the reference fields of Fig. \ref{fig:plots_ref}. These give a visual impression of the reconstruction ability of each model. In Fig. \ref{fig:plots_corr} and Fig. \ref{fig:plots_rmse}, we show the Pearson correlation and Root Mean Square Error (RMSE) for each one of the aforementioned models. The metrics in these maps are computed for each grid point independently and for the whole year of 2018 (containing 2920 holdout grids). In Table \ref{tab:metrics}, we give a more complete list of metrics (including the MAE, SSIM and peak signal-to-noise ratio (PSNR)) computed for each grid pair separately (time step) and then summarized by taking the mean and standard deviation over the samples of the holdout year.

\begin{table}[ht]
\caption{Metrics computed for each time step, downscaled product with respect to the reference grid, of the holdout year for the models showcased this section \label{tab:metrics}.}
\centering
 \begin{tabular}{c c c c c c} 
 \hline
 Panel  & MAE           & RMSE          & PearCorr      & SSIM          & PSNR      \\ [0.5ex] 
 \hline\hline
 A      & 2.58 $\pm$ 0.92 & 4.37 $\pm$ 1.50 & 0.64 $\pm$ 0.11 & 0.81 $\pm$ 0.05 & 32.95 $\pm$ 2.97 \\ 
 B      & 1.56 $\pm$ 0.54 & 2.70 $\pm$ 0.87 & 0.85 $\pm$ 0.04 & 0.89 $\pm$ 0.03 & 34.61 $\pm$ 2.78 \\
 C      & 2.58 $\pm$ 0.93 & 4.40 $\pm$ 1.47 & 0.64 $\pm$ 0.11 & 0.88 $\pm$ 0.04 & 38.85 $\pm$ 3.00 \\
 D      & 1.60 $\pm$ 0.60 & 4.75 $\pm$ 5.03 & 0.76 $\pm$ 0.20 & \textbf{0.99 $\pm$ 0.01} & \textbf{64.87 $\pm$ 6.28} \\
 E      & \textbf{1.49 $\pm$ 0.51} & \textbf{2.56 $\pm$ 0.84} & \textbf{0.88 $\pm$ 0.03} & 0.90 $\pm$ 0.03 & 35.23 $\pm$ 2.83 \\
 F      & 1.70 $\pm$ 0.58 & 2.87 $\pm$ 0.90 & 0.84 $\pm$ 0.04 & 0.87 $\pm$ 0.03 & 34.60 $\pm$ 2.81 \\
 G      & 1.53 $\pm$ 0.56 & 2.68 $\pm$ 0.96 & 0.86 $\pm$ 0.04 & 0.89 $\pm$ 0.03 & 34.97 $\pm$ 3.03 \\
 H      & 1.51 $\pm$ 0.55 & 2.64 $\pm$ 0.90 & 0.87 $\pm$ 0.03 & 0.90 $\pm$ 0.03 & 35.03 $\pm$ 2.97 \\ [1ex]
 \hline
 \end{tabular}
\end{table}

For this particular task and datasets, and according to the metrics of Table \ref{tab:metrics}, we find that models trained in MOS fashion perform better than those in PerfectProg and that a supervised model with residual blocks and subpixel convolution upsampling provides the best results. Also, we note that models trained with a LCB perform better that those without it, thanks to the fact that the LCB learns grid point- or location-specific weights. As expected, when the same network is optimized with respect to a DSSIM+MAE loss, it reaches higher scores in the SSIM and PSNR metrics. Training the CGAN model for more epochs could improve the results but would require longer runtimes and computational resources.

\section{Summary and future directions}
In this paper, we have presented \texttt{DL4DS}, a \texttt{Python} library implementing state-of-the-art and novel DL algorithms for empirical downscaling of gridded data. It is built on top of \texttt{Tensorflow/Keras} \citep{keras}, a modern and versatile DL framework. Among the key strengths of \texttt{DL4DS} are its simple API consisting of a few user-facing classes controlling the training procedure, its support for distributed GPU training (via data parallelism) powered by \texttt{Horovod} \citep{horovod}, the generous number of customizable DL-based architectures included, the possibility of composing new architectures based on the general purpose building blocks offered, the option of saving and retraining models, and its support of different downscaling frameworks that learn either with explicit low-resolution and high-resolution data (MOS-like) or only with a high-resolution dataset (PerfectProg-like). \texttt{DL4DS} is a powerful tool for reproducible empirical downscaling experiments and for network architecture optimization. A thorough ablation study varying the depth and width of the each backbone, in combination with other design choices, such as the upsampling method, and tested over several benchmark datasets, shall allow a complete comparison of CNN-based DL models for empirical downscaling. 

DL has shown promise in post-proceesing and bias correction tasks \citep{rasp18, gronquist20}, so a natural research direction is to explore the application of \texttt{DL4DS} to related post-processing problems. Other interesting research directions are the implementation of uncertainty estimation techniques besides Monte Carlo dropout, the extension of \texttt{DL4DS} models to the case of non-gridded reference data (observational station data), the implementation of alternative adversarial losses such as the Earth mover's distance (Wasserstein GAN), and the inclusion of alternative DL algorithms such as normalizing flows \citep{groenke20}. 





\begin{Backmatter}

\paragraph{Acknowledgments}
We wish to acknowledge F. J. Doblas-Reyes, K. Serradell, H. Petetin, Ll. Lledó and Ll. Palma for helpful discussions and comments, and P-A. Bretonnière and M. Samsó for assistance with data management.

\paragraph{Funding Statement}
C.A.G.G. acknowledges funding from the European Union’s Horizon 2020 research and innovation programme under the Marie Skłodowska-Curie grant agreement H2020-MSCA-COFUND-2016-754433..


\paragraph{Code Availability}
\texttt{DL4DS} code repository: \url{https://github.com/carlos-gg/dl4ds}.




\bibliographystyle{apalike}
\bibliography{dl4ds.bib}



\end{Backmatter}
\end{document}